\documentclass[letterpaper, 10 pt, conference, final]{ieeeconf}  %

\IEEEoverridecommandlockouts                              %

\usepackage{amsfonts}
\usepackage{dsfont}
\usepackage{mathtools} %
\usepackage{bm}
\usepackage{isomath} %
\usepackage{acro} %
\usepackage{booktabs} %
\usepackage{xargs}
\usepackage{subcaption}
\usepackage{xcolor}
\usepackage{xparse}
\usepackage{listings}
\usepackage{rotating}
\usepackage{multirow}
\usepackage[]{algorithm2e}
\usepackage{pifont}%
\usepackage{siunitx}
\usepackage{float}
\usepackage{multirow}

\newcommand{\briefref}{}

\newcommand{\AC}[1]{\ac[long-format=\MakeUppercase]{#1}}

\newcommand{\mat}[1]{\mathbfit{#1}} %
\DeclarePairedDelimiter{\card}{\vert}{{\vert}} %

\newcommand{\R}{\mathbb{R}}  %
\DeclarePairedDelimiter{\norm}{\lVert}{\rVert} %
\newcommand{\pnorm}[2]{\norm{#1}_{#2}} %

\newcommand{\expnmb}[2]{{#1}\mathrm{e}{#2}}

\DeclareMathOperator*{\argmin}{arg\,min}

\newcommand{\rspc}[1]{\R^{#1}}

\ifdefined\briefref

\newcommand{\figupper}{Fig.}

\newcommand{\secupper}{Sec.}

\newcommand{\eqlower}{eq.}

\else

\newcommand{\figupper}{Figure}

\newcommand{\secupper}{Section}

\newcommand{\eqlower}{equation}

\fi

\newcommand{\Figref}[1]{\figupper~\ref{#1}}

\newcommand{\Secref}[1]{\secupper~\ref{#1}}

\let\oldeqref\eqref
\renewcommand{\eqref}[1]{\eqlower~\oldeqref{#1}}  %

\newcommand{\Tabref}[1]{Table~\ref{#1}}

\DeclareAcronym{DON}
{
	short = DON ,
	long = Dense Object Nets
}

\DeclareAcronym{NTXENT}
{
	short = NT-Xent ,
	long = normalized temperature-scaled cross entropy
}

\DeclareAcronym{SLAM}
{
	short = SLAM ,
	long = simultaneous localization and mapping 
}

\DeclareAcronym{SfM}
{
	short = SfM ,
	long = structure-from-motion
}

\DeclareAcronym{AUC}
{
	short = AUC ,
	long = area under the curve
}

\newcommand{\img}[2]{\mat{I}_{#1}^{#2}}
\newcommand{\dist}[2]{D_{#2}\left(#1\right)}

\newcommand{\pckat}[1]{\mathrm{PCK@}#1}

\newcommand{\Set}[1]{\mathcal{#1}}  %

\usepackage[font=small,labelfont=small]{caption}
\title{\LARGE \bf
Efficient and Robust Training of Dense Object Nets for Multi-Object Robot Manipulation
}

\author{David B. Adrian$^{1,2}$, Andras Gabor Kupcsik$^{1}$, Markus Spies$^{1}$, Heiko Neumann$^{2}$%
	\thanks{$^{1}$ Bosch Center for Artificial Intelligence, Renningen, Germany, 
		{\tt\small firstname(s).lastname@de.bosch.com}}%
	\thanks{$^{2}$  Institute of Neural Information Processing, Ulm University, Ulm, Germany,
		{\tt\small firstname.lastname@uni-ulm.de}}%
}

\begin{document}
\maketitle
\thispagestyle{empty}
\pagestyle{empty}

\begin{abstract}
	We propose a framework for robust and efficient training of \ac{DON} \cite{florence_dense_2018} with a focus on 
	industrial multi-object robot manipulation scenarios.
	\ac{DON} is a popular approach to obtain dense, view-invariant object descriptors, which can be used for a multitude of downstream tasks in robot manipulation, such as, pose estimation, state representation for control, etc.
	However, the original work \cite{florence_dense_2018} focused training on singulated objects, with limited results on instance-specific, multi-object applications.
	Additionally, a complex data collection pipeline, including 3D reconstruction and mask annotation of each object, is required for training.
	In this paper, we further improve the efficacy of \ac{DON} with a simplified data collection and training regime, that consistently yields higher precision and enables robust tracking of keypoints with less data requirements.
	In particular, we focus on training with multi-object data instead of singulated objects, combined with a well-chosen augmentation scheme.
	We additionally propose an alternative loss formulation to the original pixelwise formulation that offers better results and is less sensitive to hyperparameters.
	Finally, we demonstrate the robustness and accuracy of our proposed framework on a real-world robotic grasping task.
\end{abstract}

\acresetall
\section{INTRODUCTION}
A well-chosen object representation is an integral part of any robotic manipulation pipeline and is the basis for downstream tasks, such as grasping or assembly. 
With object CAD models and annotated training datasets, direct pose estimation methods are readily available~\cite{hodavn2020bop}.
However, such annotations are expensive and not feasible for many industrial manipulation scenarios, where a fast and inexpensive teach-in process for new objects is of essence.
Without annotated data, proposed object representations range from hand-designed features such as SIFT \cite{lowe_distinctive_nodate, lowe_object_1999}, to implicit representations obtained through end-to-end learning of the task \cite{levine_end--end_2016, finn_deep_2016}, or self-supervised approaches such as \ac{DON} \cite{florence_dense_2018} or grasp2vec \cite{jang_grasp2vec_2018}.

\Ac{DON} \cite{florence_dense_2018} is a recently proposed self-supervised constrastive learning framework for view-invariant object representations.
The DON model provides a dense local descriptor representation by projecting a query RGB image $\img{}{} \in \rspc{H \times W \times 3}$ to a dense descriptor image $\img{D}{} \in \rspc{H \times W \times D}$, with $D\in \mathbb{N}^{+}$, the dimension of the descriptor space.
\ac{DON} exploit the geometric prior of a sequence of registered RGBD frames to sample pixel correspondences between alternative views of the same object.
The correspondence information is then used with contrastive learning to train a model that uniquely embeds the visual appearance of objects in descriptor space.
Offering intra-class generalization, consistent descriptors for deformable objects, and more, \ac{DON} have been adapted to several use-cases, such as rope manipulation \cite{sundaresan_learning_2020}, block stacking \cite{chai_multi-step_2019}, control \cite{manuelli_keypoints_2020}, and 6D grasp pose estimation \cite{kupcsik_supervised_2021}.

\begin{figure}
	\centering
	\includegraphics[width=\linewidth]{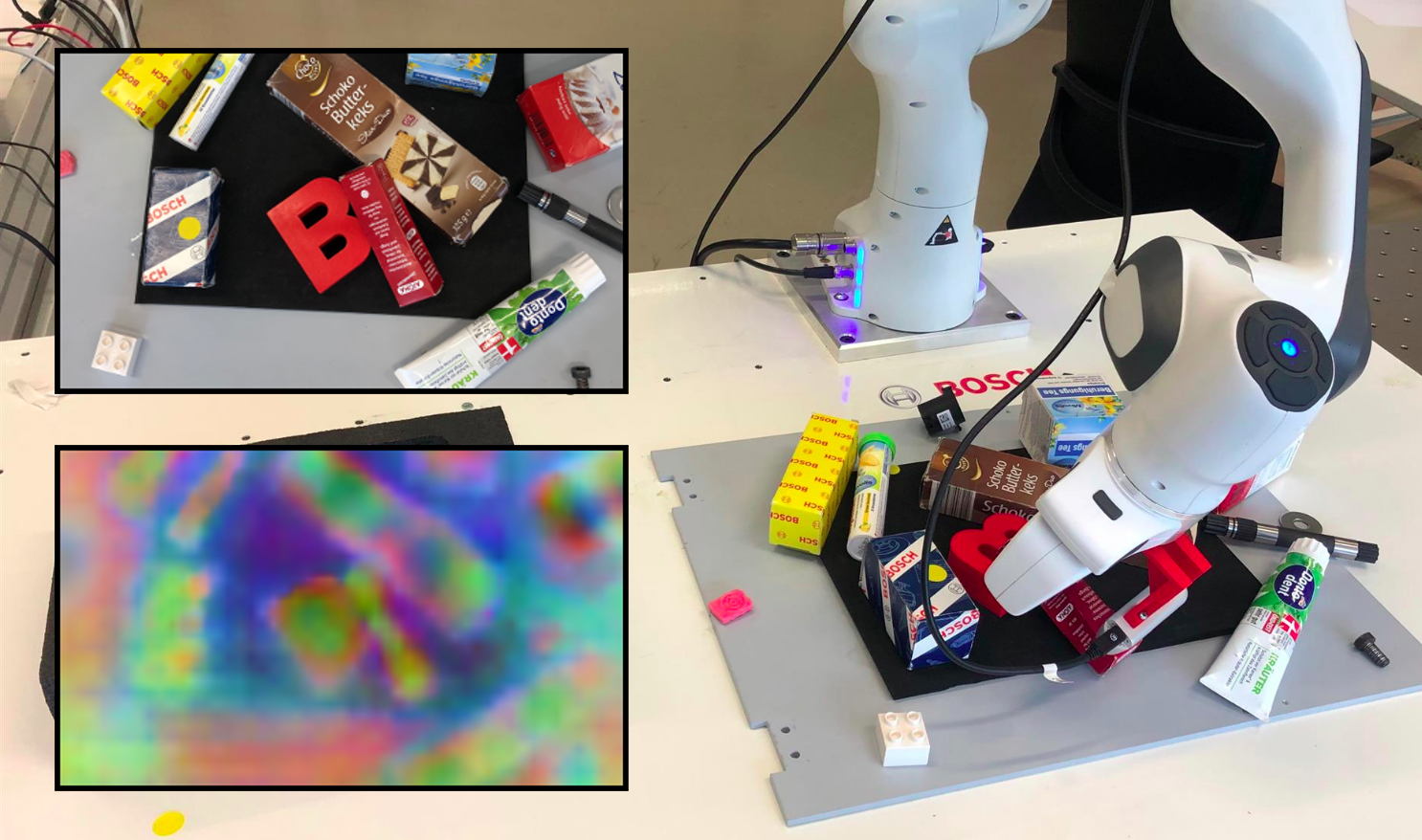}
	\caption{
		Detection and axis-oriented grasping of a target object (small, red box) in a cluttered scene.
		A typical application, where vanilla \ac{DON} fails easily, but our proposed method performs robustly.
		The image depicts our `multi-object, isolated' setting, where the target object is mostly separated from other objects, see \Secref{sec:robotic_experiment}.
		Top left: The camera perspective during \emph{object detection}.
		Bottom left: The descriptor image, visualized as a low-resolution PCA projection to 3 dimensions for illustrative purposes.
	}
	\label{fig:grasp_task}
\end{figure}

The original formulation of \ac{DON} mostly focuses on singulated objects, both during training, as well as in its demonstrated use-cases.
However, in industrial applications, e.g., grasping specific objects in a heap, often multiple objects in a potentially cluttered environment need to be detected and manipulated.
Here objects are commonly coming from a fixed and known set.
For these applications, we require distinct and easily trackable descriptors that robustly differentiate between every object of the manipulation task, instead of intra-class generalization.
Also reported by Florence et al. \cite{florence_dense_2018}, the original DON training approach is prone to fail in such multi-object scenarios, even with their proposed cross-object loss.
While \cite{florence_dense_2018} offers an early discussion on instance-specific training for multi-object scenarios, no quantitative results are provided.
Furthermore, the self-supervised training approach of \ac{DON} requires the generation of object masks, which is cumbersome to automate and error prone in practice.
This also prevents training on data with naturally occurring occlusion or the inclusion of unknown, distracting objects in the scene, which are important to create realistic training data.

In this work, we improve the training pipeline of \ac{DON} in different aspects.
First, we present a simplified data collection process for training by recording scenes of multiple objects as opposed to only one object per scene.
This allows us to record more views per object, including occlusions. 
Obtaining as many diverse perspectives per object is important to obtain view- and scale-invariant features, thus increasing our data efficiency by a large factor.
As scenes can now be densely packed, we no longer require object masks to ensure a good ratio between object and background correspondences.
This greatly reduces the technical challenges and makes the training pipeline easier to fully automate.

Second, to further improve data-efficiency, we propose a data augmentation scheme as more commonly seen in state-of-the-art visual contrastive learning \cite{chen_simple_2020, grill_bootstrap_2020, zbontar_barlow_2021, caron_unsupervised_2021}, that helps to prevent overfitting while training, but also increases accuracy and provides robustness in practice.

Finally, as loss function, 	we present an adaptation of InfoNCE \cite{oord_representation_2019}, also popularized as \ac{NTXENT} in the visual domain by SimCLR \cite{chen_simple_2020}.
As opposed to the pixelwise contrastive loss of \cite{florence_dense_2018}, we found it to be less sensitive to its hyperparameters, while offering better performance at the same time.

\section{RELATED WORK}
\label{sec:related_work}
In case of known CAD models and available annotated training datasets, pose estimation for manipulation from RGB(-D) images have recently shown great results \cite{hodavn2020bop}.
In our use-case, we focus on scenarios where such data is not available and review related work in that respect.

Here, correspondence estimation is at the core of many computer vision problems such as 3D reconstruction, \ac{SLAM} and \ac{SfM}, or object recognition and localization.
This is often achieved by (sparse) local visual feature detection and description, with methods ranging from hand-crafted, generic features such as SIFT \cite{lowe_distinctive_nodate} or ORB \cite{Rublee2011}, to learned feature-pipelines like LIFT \cite{YiTLF16}, L2-Net \cite{TianFW17}, Universal Correspondence Networks (UCN) \cite{ChoyGSC16}, SuperPoint \cite{DeToneMR18}, or KP3D \cite{tang2020kp3d}.
We also find hybrid approaches, e.g., in the domain of object pose estimation Pitteri et al.~\cite{pitteri20203d} present a hand-crafted transition- and orientation-invariant surface feature embedding that describes local geometries combined with a network trained to predict these features from RGB images.

In this work, we focus on contrastive learning of object-centric dense local visual descriptors.
Extending the dense pixelwise correspondence estimation framework in \cite{SchmidtNF17} and UCN, Florence et al. \cite{florence_dense_2018} proposed \ac{DON}, a fully automatable approach with a focus on object manipulation. It was applied to various tasks, including block stacking \cite{chai_multi-step_2019}, 6D grasp pose estimation \cite{kupcsik_supervised_2021}, control \cite{manuelli_keypoints_2020, florence_self-supervised_2019}, or rope manipulation \cite{sundaresan_learning_2020}.

Global visual descriptor learning is a closely related topic.
Here, typically a Siamese network is used to process augmentations of the same image and find underlying features that are invariant to these mappings.
SimCLR \cite{chen_simple_2020}, SwAV \cite{caron_unsupervised_2021}, or MoCo \cite{he_momentum_2020} and MoCo-v2 \cite{chen2020moco} all utilize some form of contrastive(-like) learning.
In contrast, BYOL \cite{grill_bootstrap_2020} only uses positive pairs and Barlow Twins \cite{zbontar_barlow_2021} optimizes the empirical cross-correlation.
Similar ideas were recently also adapted for general local descriptor learning, e.g., in DenseCL \cite{wang_dense_nodate} or combined global-local descriptor learning \cite{chaitanya_contrastive_nodate}.

Over the last years many impressive object manipulation solutions emerged in the robot learning community, typically relying on inexpensive RGB(-D) cameras. 
For example, DexNet-based algorithms were proposed for grasping using depth images \cite{Mahler2017}.
Zeng et al. use vision-guided learning to grasp unknown objects for a throwing task \cite{Zeng2019}.
Pinto and Gupta investigate large-scale self-supervision  for grasp learning \cite{Pinto2016}, while Kalashnikov et al. \cite{Kalashnikov2018} employ reinforcement learning at large-scale to grasp from a heap. One of the closest works to \ac{DON} for object manipulation is proposed by Vecerik et al. \cite{Vecerik2020}.
They show how task-relevant object keypoints can be learned by human- and self-supervision, e.g., to learn cable plugging.

\section{METHOD}
\label{sec:method}
Similar to \cite{florence_dense_2018}, we record static scenes including objects we wish to learn using a robot and a wrist-mounted RGBD camera.
Every recorded scene consists of thousands of registered RGBD images. 
As in the original DON approach, we randomly sample image pairs $(\img{}{A}, \img{}{B})$ from unique scenes for training.
The sampled image pairs present two distinct, but partially overlapping views of the scene.

\begin{figure}
	\centering
	\begin{subfigure}[b]{0.49\linewidth}
		\centering
		\includegraphics[width=\textwidth]{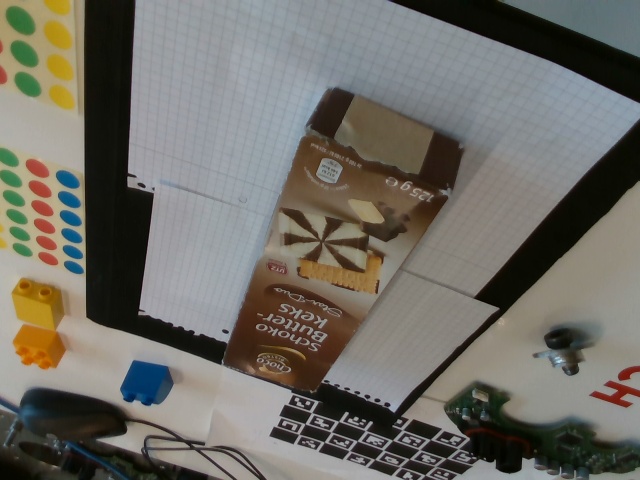}
		\caption{Single-object scene}
		\label{fig:single_object}
	\end{subfigure}
	\begin{subfigure}[b]{0.49\linewidth}
		\centering
		\includegraphics[width=\textwidth]{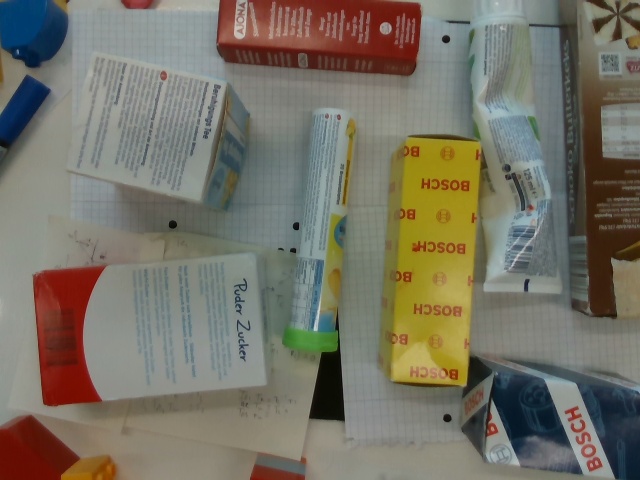}
		\caption{Multi-object scene}
		\label{fig:multi_object}
	\end{subfigure}
		\caption{Fig. (a) shows an example from the single object training data.
		Fig. (b) depicts the set of known objects, together with background clutter, from the multi-object scenes.
		Object features include repeating texture patterns, high specularity, or deformability.}
	\label{fig:recorded_dataset}
\end{figure}

\begin{figure}[h]
	\centering
	\includegraphics[trim={1.3cm -0.1cm 0.6cm 0},width=0.8\linewidth]{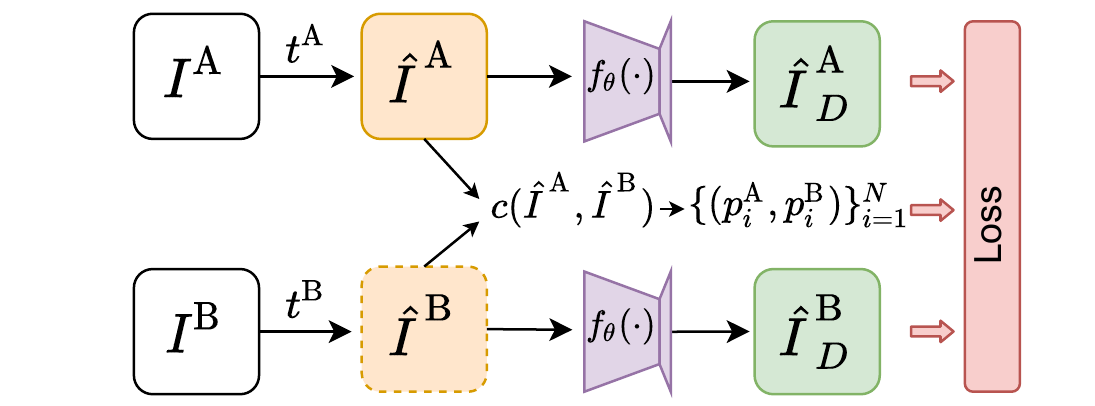}
	\caption{
		In our proposed framework image $\img{}{A}$, and optionally $\img{}{B}$, are augmented with randomly sampled sequences of augmentations $t^{\mathrm{A}}$, $t^{\mathrm{B}}$.
		Using the shared network $f_\theta$, we obtain the descriptor images.
		$c(\cdot, \cdot)$ denotes the correspondence sampling process, which returns a set of pixel correspondences used by the loss.%
	}
	\label{fig:data_flow}
\end{figure}

In our work, we record scenes that contain multiple objects we wish to learn (Fig. \ref{fig:multi_object}), as opposed to the single object scenes (Fig. \ref{fig:single_object}) in \cite{florence_dense_2018}.
We also skip the object mask generation step, which is, in practice, error-prone and difficult to automate.
More details about the data collection are presented in \Secref{sec:data_collection}.
In what follows we describe our training approach relying on image pairs sampled from multi-object scenes.
Specifically, we focus on image augmentation, pixel correspondence sampling, and the loss function.
We provide an overview of the general data flow in \Figref{fig:data_flow}.

\subsection{Augmentations}
\label{sec:augmentations}
Augmentations have been shown in self-supervised and contrastive learning to not only be beneficial, but integral to learning discriminative global feature representations, cf. \cite{chen_simple_2020, caron_unsupervised_2021, zbontar_barlow_2021}.
These methods typically start with a single image and create two corresponding views using two randomly sampled sequences of augmentations.
In our work, the two different views are readily sampled from the dataset, however, image augmentations can still diversify our training dataset to improve data-efficiency and mitigate overfitting.
The idea of augmentations for \ac{DON} was also used in \cite{florence_dense_2018} with background randomization and vertical/horizontal flipping.
For this study, we focus on a broader set of common augmentations: \textit{resize and crop}, \textit{perspective and affine distortions}, \textit{horizontal and vertical flips}, \textit{rotations},  \textit{blur}, \textit{color jitter}, and \textit{grayscale conversion}.

In particular, transformations such as perspective distortions, naturally appear in robotic manipulation scenarios. %
Similarly, blurring and color distortions frequently occur in practice due to changing light conditions or motion blur.
Thus, utilizing this set of augmentations should not only help to prevent over-fitting due to limited training dataset size, but also provides additional cases for improving robustness.

For both images we randomly sample a subset of augmentations and apply them (Fig.~\ref{fig:data_flow}).
In practice, however, we found that applying the transformations to only one of the images to perform better.
As \ac{DON} rely heavily of color cues, using color affecting augmentations on both images impacted training heavily.
An ablation of the individual augmentations and further discussion is summarized in \Secref{sec:ablation_augmentations}.
We utilize the readily available implementations by the Torchvision library of the Pytorch project \cite{paszke_pytorch_2019}.

\subsection{Correspondence Sampling}
\label{sec:correspondence_sampling}
Correspondence sampling for an image pair is straight-forward given the respective camera poses, intrinsic matrix, and depth information using knowledge from multiple view geometry \cite{hartley_multiple_2003}.
However, as we rely on densely packed scenes and image pairs with potentially large perspective changes, we face occlusion and partially overlapping fields-of-view.
Hence, instead of sampling individual pixels directly and checking for their validity afterwards, we propose the following direct approach: map every pixel of $\img{}{A}$ via their world-coordinate representation to the other view, then determine which pixels are actually inside the other view and not occluded.
This yields a Boolean mask $\mat{M}$, determining which pixels in $\img{}{A}$ have a true correspondence in $\img{}{B}$.
We can now randomly sample valid pixel correspondences according to $\mat{M}$ given the previously computed mapping.
We refer to these as \emph{matches}, or positive pairs, in the following.

\subsection{Loss}
\label{sec:losses}
Although the pixelwise formulation in \cite{florence_dense_2018} shows impressive results, we found it to be sensitive to its hyper-parameters.
To this end, we propose an adaptation of InfoNCE \cite{oord_representation_2019}, also called \ac{NTXENT} in recent work such as SimCLR \cite{chen_simple_2020}, for the purpose of training \ac{DON}.

Instead of the original pixelwise formulation in \cite{florence_dense_2018}, and its cross-object variant, we require only a single loss function, that works for both singulated and multi-object settings.
Here, we randomly sample $N$ correspondences from an image pair $(\img{}{\mathrm{A}},\img{}{\mathrm{B}})$, with each correspondence yielding two descriptors $d_i^{\mathrm{A}}$ and $d_i^{\mathrm{B}}$, for a total of $N$ pairs, or $2N$ descriptors.
For a given descriptor, we treat all other $2(N-1)$ descriptors respectively as negative examples.
Hence, unlike the vanilla pixelwise contrastive formulation in \cite{florence_dense_2018}, we directly optimize all descriptors with respect to each other.

For one descriptor pair $(d_i, d_j)$, we define the loss as
\begin{equation}
l_{i,j} = -\log \frac{\exp(\dist{d_i, d_j}{}/\tau)}{\sum_{k=1; k\neq i}^{2N}  \exp(\dist{d_i, d_k}{}/\tau)},
\label{eq:ntxent}
\end{equation}
where $\tau$ is a temperature scaling factor and  $\dist{\cdot,\cdot}{}$ denotes a distance or a similarity measure.
The complete loss is given by the sum over all pairwise losses $l_{i,j}$.
This loss is comparable to the one in \cite{chen_simple_2020}, where each image pair in a mini-batch is considered as negatives.
In our work instead, we compute the loss  directly on pixel-level descriptors.
Finally, we coalesce all correspondences across the image pairs in the batch into a single mini-batch dimension.
In the following, we use the \textit{cosine similarity}, defined as
\begin{equation}
	\dist{d_i, d_j}{} =
		\frac{
			\langle d_i, d_j \rangle
		}{
			\pnorm{d_i}{2} \pnorm{d_j}{2}
		},
\end{equation}
that is the dot product between vectors that have been normalized to unit length.

\section{EVALUATION}
\label{sec:evaluation}

In this section we compare our proposed training pipeline to that of vanilla DON.
We first describe how data collection and training are performed.
Then, we explain our metrics for two different evaluation scenarios: i) we show how accurately the different models predict correspondences in image plane, and ii) for downstream tasks we measure keypoint tracking accuracy in world coordinates.
Finally, we perform ablation studies and show results on a real world robot grasp experiment.

\subsection{Hardware Setup}
We used a Franka Emika Panda 7-DoF arm, equipped with a wrist-mounted Intel RealSense D435 camera for recording of RGB-D images, see \Figref{fig:grasp_task}.
All scenes were recorded using the same robot end-effector trajectory, at $640 \times 480$px and $30\si{\Hz}$.
We sub-sampled recordings by enforcing at least $5^{\circ}$ rotation and $2\si{cm}$ translation between consecutive frames, yielding around 450 frames per scene.
A parallel gripper was employed for the grasping tasks.

\subsection{Data Collection}
\label{sec:data_collection}
We selected a set of 8 objects for training, including objects with repeating patterns, similar visual features, and one deformable object.
We collected separate sets of training data for the multi-object and single-object setting, see \Figref{fig:recorded_dataset}.
For \emph{single-object} training with vanilla and cross-object DON \cite{florence_dense_2018} we recorded multiple scenes for each object isolated in the workspace, for a total of 20 scenes.
For \textit{multi-object} data, we captured 15 scenes in total, each containing a subset of the known objects.
For validation we recorded one multi-object and 5 single-object scenes, each with some unknown objects in the vicinity.
A total of 8 multi-object scenes were recorded for the test set.
In each test scene, a subset of the 8 objects were presented in different configurations, with some additional unseen objects that serve as distractions. %
We manually provided object-specific masks for each of the known objects, allowing us to sample query points specifically for each objects for validation and testing.

\subsection{Training}
\label{sec:training}
Following \cite{florence_dense_2018}, and for comparison with existing literature, we utilized the same ImageNet-pretrained ResNet-34 with output stride of 8, that is upsampled to the original input dimension; cf. the public reference implementation of \cite{florence_dense_2018}.
We note that, unlike various state-of-the-art models \cite{grill_bootstrap_2020, chen_simple_2020,caron_unsupervised_2021}, we do not employ a projection layer, or other advanced mechanisms like gradient stopping, after extracting the representation, but directly apply the loss as in \cite{florence_dense_2018}.

We trained using ADAM \cite{kingma_adam_2015} optimizer, with a learning rate of $\expnmb{3}{-5}$, momentum of $0.9$, and weight decay of $\expnmb{1}{-4}$.
All models were trained for 500 epochs, with each epoch consisting of 1000 image pairs.
Our approach using \ac{NTXENT} loss was trained with a batch size of 2, while the pixelwise contrastive losses using a batch size of 1 (as in \cite{florence_dense_2018}), as we found this to perform best, respectively, see \Secref{sec:batchsize}.
We performed validation every 5 epochs and retained the checkpoint with the best validation results with respect to the $\pckat{80}$ metric, see \Secref{sec:metrics} for its definition.

For \ac{NTXENT} we sampled 2048 correspondences per image, yielding a total of 4096 correspondence pairs for a batch size of 2.
Pixelwise contrastive losses were trained using 20,000 correspondences for an image pair with each 150 non-correspondences per match.
This is different to the 10,000 match used in \cite{florence_dense_2018}, as we found this to considerably increase performance, see \Secref{sec:ablation_n_correspondences}.
In the case of pixelwise cross-object loss proposed in \cite{florence_dense_2018}, we always sampled a triplet of images, that is, two corresponding views and one non-corresponding view of a different object.
This differs from \cite{florence_dense_2018}, where 50\% of training iterations are performed using either regular or cross-object non-match loss.
We found that we obtained the same or better results with our combined loss.

All models are trained by default with a descriptor dimension of $D=16$.
We trained each model configuration 5 times, each with a random seed, and reported the averaged results for each of the following experiments, where $\sigma$, if given, denotes the standard deviation.

\subsection{Evaluation Metrics}
\label{sec:metrics}
To measure performance we adopted the $\pckat{k}$ metric, as in preceding works \cite{florence_dense_2018, chai_multi-step_2019}, defined for a set of pixel pairs $\Set{T}=\{p_i, q_i\}_{i=1}^N$ as
\begin{equation}
\pckat{k}(\Set{T}) =
	\frac{1}{\card{\Set{T}}} \sum_{(p_i, q_i)\in\Set{T}} \left[ \pnorm{p_i - q_i}{2} \le k \right],
\end{equation}
where $k$ is the maximum allowed error, measured in pixels, for a prediction to still be counted as correct. $\Set{T}$ is a set of predicted $p_i$ and ground-truth $q_i$ pixels, and $[\cdot]$ denotes the \textit{Iverson bracket}, which evaluates to 1 if the logical proposition inside is satisfied and 0 otherwise.

We evaluated the function many times over a range, that is $\forall k\in\left[1, 100\right]$, and report the \ac{AUC}.
We do not consider $k>100$ as such a mismatch in image plane, with a diagonal of 800 pixels, leads to failure in downstream tasks.
Additionally, in \Secref{sec:kp_tracking}, we evaluated the model accuracy via a keypoint tracking task, measuring world-coordinate prediction errors in millimeters, which is relevant for downstream tasks.

In both cases, we selected $N$ query pixel locations $\{q_i^k\}_{i=1}^N$ on an object in a given image $\img{}{k}$.
Subsequently, the descriptor representation at these pixel locations $\{f_\theta(\img{}{k})_{q_i}\}_{i=1}^{N}=\{d_i^{k}\}_{i=1}^N$ are extracted.
We found the closest matching descriptor for each query point in some other view $l$ via its descriptor image $\img{D}{l}$ as prediction $p_i^* = (u_i^*, v_i^*) = \argmin_{u_i, v_i} \dist{\img{D}{l}(u_i, v_i),\, d_i^k}{}, \forall i\in\left[1, N\right]$.
Here, $D$ denotes the cosine similarity for \ac{NTXENT} variants, or the $\ell_2$ distance for pixelwise losses.

\subsection{Results}
\label{sec:results}

\subsubsection{Correspondence Accuracy}
\label{sec:corrs_accuracy}
As first evaluation, we selected 500 image pairs per object type across all scenes, utilizing the object-specific masks from our multi-object test data.
We randomly sampled given the constraints of a minimum translation of 20\si{cm} and minimum angle of $\pi/12$ between poses, ensuring enough variation in perspective to test for view-invariance.
Per image pair 100 query points are randomly selected on the object in one of the images.
Given these query descriptors, we found their respective closest matching descriptor in the second image as described above and calculated the $\pckat{k}$ errors.
The sampled image pairs and correspondences were evaluated for each model configuration.

\Tabref{tbl:pck_results} summarizes the performances of each loss variation.
Clearly, multi-object trained models outperform both vanilla and cross-object loss trained \ac{DON}.
For vanilla pixelwise, and its cross-object variant, augmentations are detrimental.
We partially attribute this to augmentations, such as perspective distortion, that reduced the already small object area even more.
Consequently, masked sampling starts to become inadequate for training efficiently, requiring more epochs.

\begin{table}%
	\centering	%
	\begin{tabular}{l|c}
				& \ac{AUC}$\pm \sigma$ for $\pckat{k}$, \\
		   & $\forall k \in \left[1, 100\right]$\\
		\midrule
		Pixelwise				 					& $0.313 \pm 0.011$ \\
		Pixelwise + Augmentations		 			& $0.279 \pm 0.019$ \\
		Pixelwise + Cross-Object 					& $0.387 \pm 0.017$ \\
		Pixelwise + Cross-Object  + Augmentations 	& $0.290 \pm 0.036$ \\
		Pixelwise + Multi-Object + Augmentations 	& $0.565 \pm 0.008$ \\
		\midrule
		\AC{NTXENT} + Multi-Object 					& $0.514 \pm 0.027$ \\
		\textbf{\AC{NTXENT}  + Multi-Object + Augmentations} & $\mathbf{0.621 \pm 0.015}$ \\
		\bottomrule
	\end{tabular}
	\caption{Correspondence accuracy over randomly sampled image pairs per loss and training configuration.
	}
	\label{tbl:pck_results}
\end{table}

\subsubsection{Keypoint Tracking}
\label{sec:kp_tracking}
Using dense descriptor representations for downstream tasks requires detecting keypoints relevant for the task.
The keypoints can be defined manually, or by a selection algorithm.
Thus, in the following we evaluated how accurately we can track keypoints in world coordinates using the different models.

\begin{figure}[h]
	\centering
	\includegraphics[width=\linewidth]{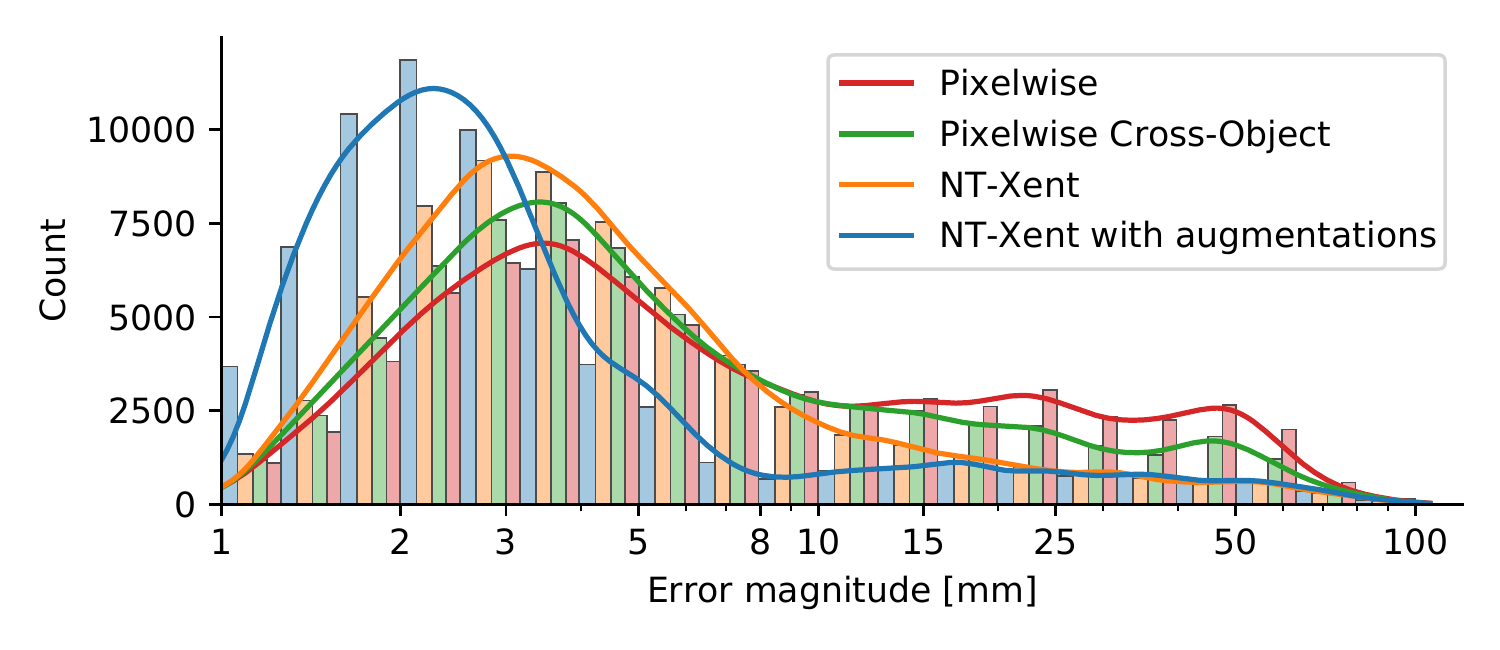}
	\caption{
		Error histograms for the keypoint tracking task, reported in millimeter on a log-scale.
		Lines are the kernel density estimate.
	}
	\label{fig:wc_error_plot}
\end{figure}

We treated our recorded scenes as video-sequences with length $M$.
We manually annotated query points $\{q_i^k\}_{i=1}^N$ on a single frame, which serve as reference keypoints to track. %
Selecting between $2-5$ query points per object and sequence, we computed predictions for each query point in the remaining $M-1$ images.
After detecting the closest pixels to the queries in every frame, as described in \Secref{sec:metrics}, we projected the corresponding depth value to world-coordinates.
Note that we have access to the camera intrinsic and extrinsic parameters, therefore this computation is consistent within a sequence.
Finally, the error is calculated as the Euclidean distance between the respective world-coordinate representations $\pnorm{\tilde{p}_i^{*} - \tilde{q}_i^k}{2},~\tilde{p},\tilde{q}\in \mathbb{R}^3$.
The prediction error histograms averaged over all queries are shown in Fig. \ref{fig:wc_error_plot}.

All models exhibit clear long tail distributions.
However, our training pipeline with \ac{NTXENT} and augmented multi-object scenes led to more accurate and robust tracking performance.

\subsection{Ablation}
\label{sec:ablation}

All following results are reported with respect to the correspondence accuracy task in \Secref{sec:corrs_accuracy}.
Models are with default parameters, with one aspect varied as described.
Each configuration is trained 5 times, and averaged results are reported.

\subsubsection{Data Efficiency}
\label{sec:single_vs_multi}
\begin{figure}[h]
	\centering
	\includegraphics[width=\linewidth]{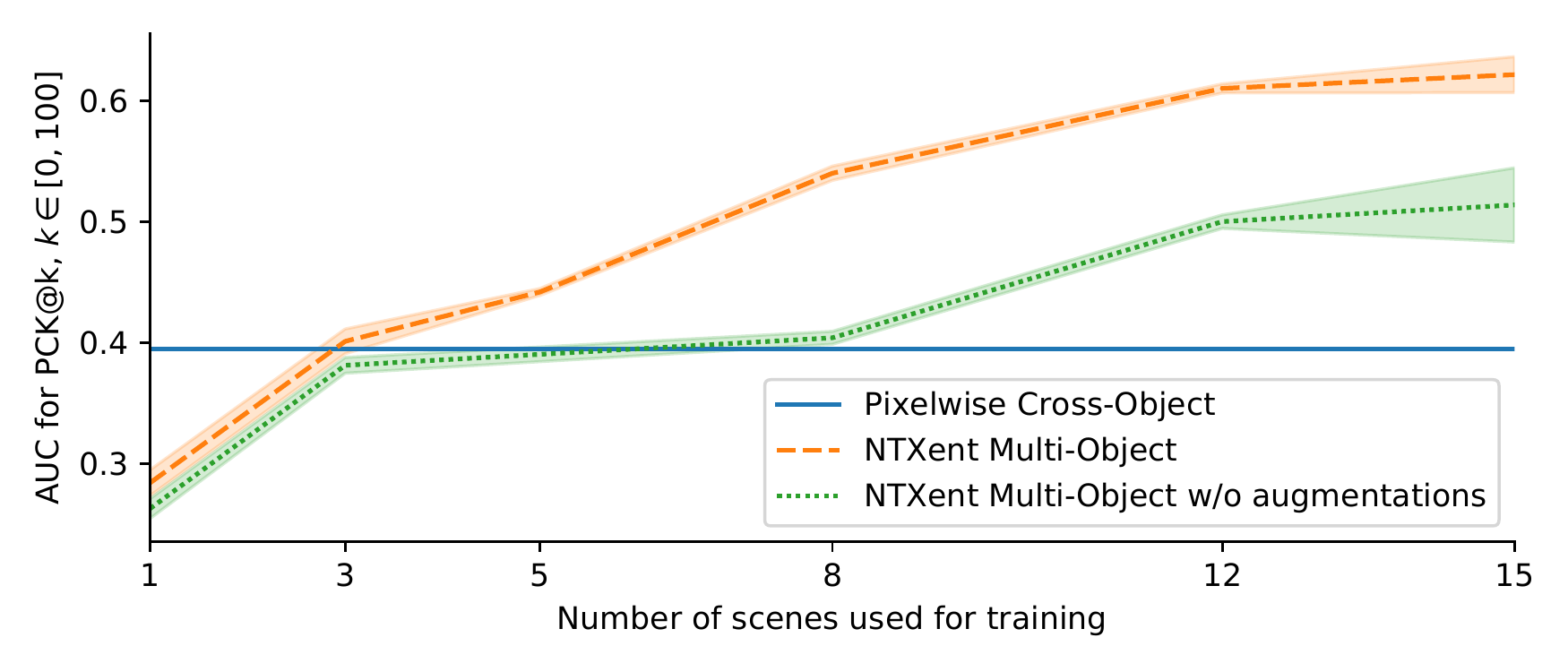}
	\caption{
		Contribution of successively adding more scenes to the multi-object training dataset.
		Pixelwise cross-object, plotted as reference line, was trained with all 20 singulated-object scenes.
	}
	\label{fig:ablation_data}
\end{figure}

We evaluated the impact of available data with respect to the overall performance in comparison to training on  singulated objects.
We found a much higher data efficiency, see \Figref{fig:ablation_data}, as with just 3 multi-object scenes and augmentations, we obtained the same performance as the pixelwise cross-object loss with 20 scenes.
Adding further scenes increases performance, with augmentations adding a larger benefit when not much data is available.

\subsubsection{Augmentations}
\label{sec:ablation_augmentations}
\begin{figure}[h]
	\centering
	\includegraphics[width=\linewidth]{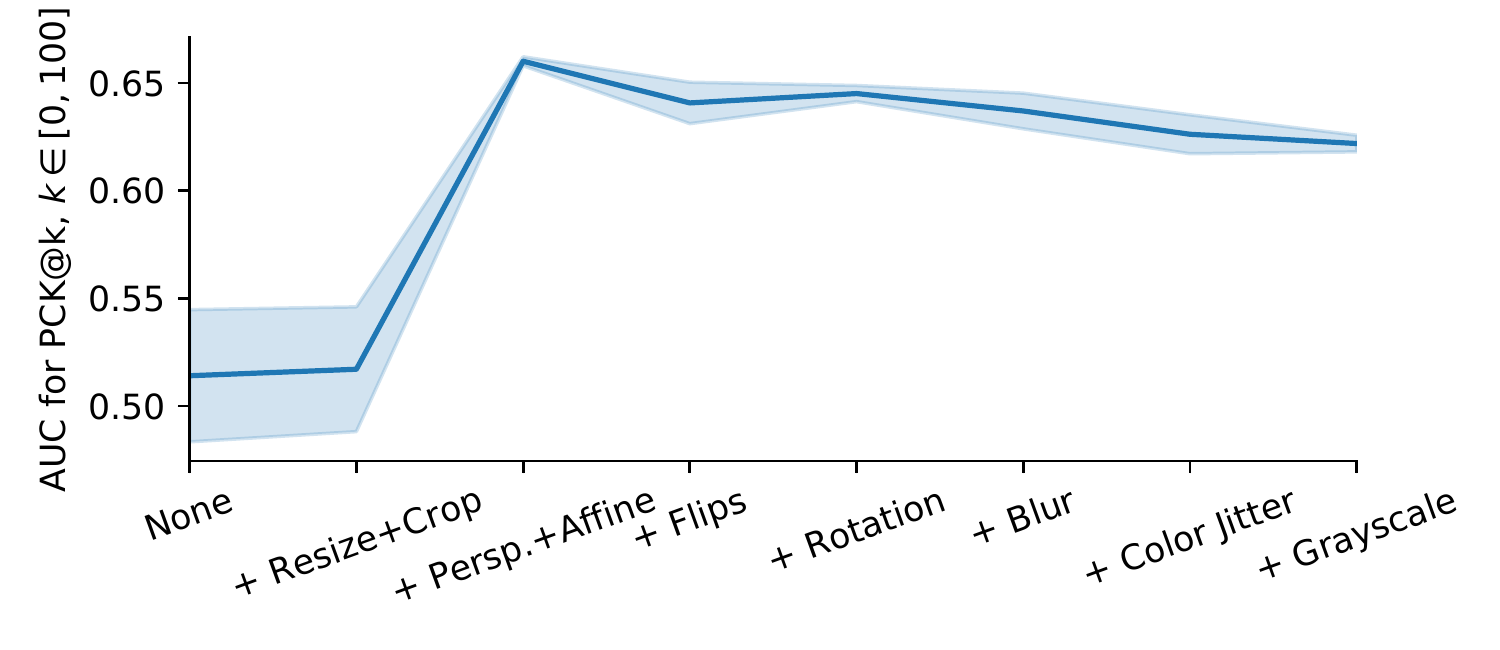}
	\caption{
		Contribution of successively adding augmentations to the training.
		Standard deviation plotted as error band.
	}
	\label{fig:ablation_augmentation}
\end{figure}

In order to evaluate which augmentations are beneficial, and how much they respectively contribute to the overall performance, we trained multiple configurations of the \ac{NTXENT} multi-object model starting with no augmentations and each time incrementally adding new augmentations, see \Figref{fig:ablation_augmentation}.

The most beneficial augmentations proved to be perspective + affine distortions, which both imitate the naturally occurring rotations and perspective distortion in the multi-view training data.
Not surprisingly, applying further augmentations like vertical/horizontal flipping and rotations do not add any more benefits.

Augmentations affecting the color distribution of the image, that is, \textit{color jitter} and \textit{grayscale}, led to small decrease in accuracy.
This indicates that, at least on our data, \ac{DON} might exploit accurate color cues, instead of generalizing to other features.
On the keypoint tracking task, the differences are negligible as the physical error increased by less than 0.5\si{mm} for 75\% off all measured errors when adding color-affecting augmentations.

\subsubsection{Feature Dimensions}
\label{sec:ablation_dim}
Both losses profit from increasing feature dimensions, with diminishing benefits for $D>32$, cf. \Tabref{tbl:abl_dims}.

\begin{table}%
	\centering	%
	\begin{tabular}{l|ccccc}
		\toprule
		{} & \multicolumn{5}{c}{\ac{AUC} for $\pckat{k},\forall k \in\left[1, 100\right]$}\\
		$D$ & 3 & 8 & 16 & 32 & 64 \\
		\midrule
		\ac{NTXENT} & 0.318 & 0.550 & 0.621 & 0.620 & \textbf{0.634} \\
		Pixelwise 	& 0.406 & 0.532	& 0.565 & 0.593 & \textbf{0.610} \\
		\bottomrule
	\end{tabular}
	\caption{\ac{AUC} for varying latent dimensions $D$.
		Both losses are trained on multi-object data and with full set of augmentations.
	}
	\label{tbl:abl_dims}
\end{table}

\subsubsection{Temperature}
\label{sec:ablation_ntxent_temperature}
We found \ac{NTXENT} stable with values in the range $0.01-0.3$ and optimal value around $0.1$.
The performance dropped for larger values around $0.7-1.0$.
This is in line with results reported for SimCLR \cite{chen_simple_2020}.

\subsubsection{Batch Size}
\label{sec:batchsize}
\ac{NTXENT} loss is mostly invariant to the batch size, see \Tabref{tbl:batch_size}.
In contrast, the pixelwise formulation exhibits strong decrease in performance for batch-sizes larger than 1.
We hypothesize, that if each scene can only contain a small subset of objects, training with higher batch sizes could optimize more objects representations simultaneously, but this requires further investigation.

\begin{table}%
	\centering	%
	\begin{tabular}{l|ccc}
		\toprule
		{} & \multicolumn{3}{c}{\ac{AUC}$\pm \sigma$ for $\pckat{k},\forall k \in\left[1, 100\right]$}\\
		Batch Size & 1 & 2 & 4 \\
		\midrule
		Pixelwise   & $\mathbf{0.565 \pm 0.008}$ & $0.377 \pm 0.032$ & $0.310 \pm 0.017$ \\
		\ac{NTXENT} & $0.579 \pm 0.018$ & $\mathbf{0.621 \pm 0.015}$ & $0.600 \pm 0.010$ \\
		\bottomrule
	\end{tabular}
	\caption{
		\ac{AUC} for varying batch sizes.
		Both losses are trained on multi-object data and with full set of augmentations.
	}
	\label{tbl:batch_size}
\end{table}

\subsubsection{Number of correspondences}
\label{sec:ablation_n_correspondences}
Both losses appeared robust to the number of correspondences, see \Tabref{tbl:abl_correspondeces}, but pixelwise loss performance deteriorated below 20,000 correspondences.
We note that the large number of correspondences for the pixelwise loss are only possible because we do not sample exclusively within masks, which would typically only occupy around 4,000-10,000 pixels.

\begin{table}%
	\centering	%
	\begin{tabular}{cr|c}
		\toprule
		{} & \# Correspondences & \multicolumn{1}{c}{\ac{AUC}$\pm \sigma$ for $\pckat{k},\forall k \in\left[1, 100\right]$}\\
		\midrule
		\parbox[t]{2mm}{\multirow{7}{*}{\rotatebox[origin=c]{90}{Pixelwise}}}  	& 10k 						& $0.472 \pm 0.016$ \\
	 																			& 20k 						& $0.565 \pm 0.008$ \\
	 																			& 20k, \,~300 NC 			& $0.568 \pm 0.008$ \\
	 																			& \textbf{20k, 1000 NC} 	& $\mathbf{0.575 \pm 0.010}$ \\
	 																			& 30k 						& $0.564 \pm 0.007$ \\
	 																			& 40k 						& $0.558 \pm 0.014$ \\
		\midrule
		\parbox[t]{2mm}{\multirow{4}{*}{\rotatebox[origin=c]{90}{\ac{NTXENT}}}}	& 512 				& $0.625 \pm 0.004$\\
																				& \textbf{1024} 	& $\mathbf{0.627 \pm 0.012}$ \\
																				& 2048 				& $0.621 \pm 0.015$ \\
																				& 4096 				& $0.609 \pm 0.013$ \\
		\bottomrule
	\end{tabular}
	\caption{\ac{AUC} for varying number of correspondences sampled per image.
	 NC denotes the number of non correspondences per correspondence for pixelwise loss.}
	\label{tbl:abl_correspondeces}
\end{table}

\subsection{Robotic Experiment: Grasp prediction}
\label{sec:robotic_experiment}
Instead of the single point grasping shown in \cite{florence_dense_2018}, we evaluated the models on a 6D grasp pose prediction task, similar to \cite{kupcsik_supervised_2021}.
To this end, we took a single RGB image of each object, we manually selected two pixels to denote an axis, and extracted the respective descriptors $d_1, d_2$.
In the test scenes, we find the respective predictions for each descriptor and calculate their world-coordinate representations $\tilde{p_1},\tilde{p_2}$. 
Using these keypoint coordinates we infer the predicted graps position and orientation as in \cite{kupcsik_supervised_2021}.

We recorded three settings of difficulty based on how objects appear within the workspace of the robot and the view of the wrist-mounted camera:
\begin{enumerate}
	\item \textit{Singulated}: each object is presented by itself
	\item \textit{Multi-Object, Isolated}: all objects, plus additional unknown ones, are present, but each object is spatially isolated
	\item \textit{Multi-Object, Packed}: all objects, plus additional unknown ones, are present, and objects are densly packed
\end{enumerate}
For each level of difficulty, we arranged the scene such that the target object can be detected and grasped successfully.
\Figref{fig:grasp_task} depicts our setup for the multi-object, isolated scenario.

We performed three grasp attempts for every object we used for training, rotating the scene by 45 degrees for each attempt to probe for robustness.
The respective setup and orientation was kept identical for both networks.
In total, we executed 24 grasp attempts per setting and model.
The results are summarized in \Tabref{tbl:grasping_results}.
In all tasks our method demonstrated higher accuracy, more than doubling the success rate on multi-object settings.

Notably, in the singulated setting, our models' performance dropped in comparison to the harder \textit{multi-object isolated}.
However, this is not surprising as it was not trained on data of singulated objects, indicating that it overfits to the statistics of multi-object images.
This encourages further improvements to the training to mitigate this limitation.

\begin{table}%
	\centering
	\begin{tabular}{l|ccc}
		\toprule
								& Pixelwise Cross-Object &  \textbf{Ours}  \\
		\midrule
		Singulated              &  70.8\% 	&  \textbf{83.3\%} \\
		Multi-Object, Isolated	&  41.7\%	&  \textbf{95.8\%} \\
		Multi-Object, Packed 	&  25.0\%	&  \textbf{70.8\%} \\
		\midrule
		Overall 				& 45.8\%    &  \textbf{83.3\%} \\
		\bottomrule
	\end{tabular}
	\caption{Combined detection and grasp success rate per difficulty level and aggregated over all settings.}
	\label{tbl:grasping_results}
\end{table}

\section{CONCLUSION}
We presented a training framework for Dense Object Nets with a particular focus on multi-object tasks.
Our proposal reduced the required data amount and drastically simplified the data pre-processing.
We removed the need for mask generation, while improving performance for multi-object scenarios.
We investigated which image augmentations are required to improve robustness and demonstrated that our combined proposal leads to significant improvements, both on benchmark tests, as well as for oriented grasping requiring high precision.
Despite the improvements, we found that all trained \acp{DON} suffer from becoming unstable in very densely packed scenes.
We believe this to be due to network backbone used, that uses a simple bilinear upsampling to produce the dense descriptor image.
Hence, further research should investigate and focus on improving the stability of descriptors.

\bibliographystyle{IEEEtran}
\bibliography{IEEEabrv,DON}

\end{document}